%% file: conference_101719.tex
\documentclass[conference]{IEEEtran}
\IEEEoverridecommandlockouts
\usepackage{cite}
\usepackage{amsmath,amssymb,amsfonts}
\usepackage{algorithmic}
\usepackage{graphicx}
\usepackage{textcomp}
\usepackage{xcolor}
\usepackage{url}
\def\BibTeX{{\rm B\kern-.05em{\sc i\kern-.025em b}\kern-.08em
    T\kern-.1667em\lower.7ex\hbox{E}\kern-.125emX}}
\begin{document}

\title{A Framework for Human-Reason-Aligned Trajectory Evaluation in Automated Vehicles}

\author{\IEEEauthorblockN{Lucas Elbert Suryana}
\IEEEauthorblockA{\textit{Department of Transport and Planning} \\
\textit{Centre for Meaningful Human Control}\\
\textit{Delft University of Technology}\\
Delft, the Netherlands \\
L.E.Suryana@tudelft.nl}
*Corresponding author
~\\
\and
\IEEEauthorblockN{Saeed Rahmani}
\IEEEauthorblockA{\textit{Department of Transport and Planning} \\
\textit{Delft University of Technology}\\
Delft, the Netherlands \\
S.Rahmani@tudelft.nl}
~\\
\and
\IEEEauthorblockN{Simeon C. Calvert}
\IEEEauthorblockA{\textit{Department of Transport and Planning} \\
\textit{Centre for Meaningful Human Control}\\
\textit{Delft University of Technology}\\
Delft, the Netherlands \\
S.C.Calvert@tudelft.nl}
~\\
\and
\begin{minipage}[t]{0.05\textwidth}
~  % Just a non-breaking space
\end{minipage}
\and
\IEEEauthorblockN{}
\IEEEauthorblockA{}
\and
\IEEEauthorblockN{Arkady Zgonnikov}
\IEEEauthorblockA{\textit{Department of Cognitive Robotics} \\
\textit{Centre for Meaningful Human Control}\\
\textit{Delft University of Technology}\\
Delft, the Netherlands \\
A.Zgonnikov@tudelft.nl}

\and

\IEEEauthorblockN{Bart van Arem}
\IEEEauthorblockA{\textit{Department of Transport and Planning} \\
\textit{Delft University of Technology}\\
Delft, the Netherlands \\
B.vanArem@tudelft.nl}

\and
\IEEEauthorblockN{}
\IEEEauthorblockA{}
}

\maketitle

\begin{abstract}
One major challenge for the adoption and acceptance of automated vehicles (AVs) is ensuring that they can make sound decisions in everyday situations that involve ethical tension. Much attention has focused on rare, high-stakes dilemmas such as trolley problems. Yet similar conflicts arise in routine driving when human considerations, such as legality, efficiency, and comfort, come into conflict. Current AV planning systems typically rely on rigid rules, which struggle to balance these competing considerations and often lead to behaviour that misaligns with human expectations. This paper introduces a reasons-based trajectory evaluation framework that operationalises the tracking condition of Meaningful Human Control (MHC). The framework represents human agents’ reasons (e.g., regulatory compliance) as quantifiable functions and evaluates how well candidate trajectories align with them. It assigns adjustable weights to agent priorities and includes a balance function to discourage excluding any agent. To demonstrate the approach, we use a real-world-inspired overtaking scenario, which highlights tensions between compliance, efficiency, and comfort. Our results show that different trajectories emerge as preferable depending on how agents’ reasons are weighted, and small shifts in priorities can lead to discrete changes in the selected action. This demonstrates that everyday ethical decisions in AV driving are highly sensitive to the weights assigned to the reasons of different human agents.
\end{abstract}


\section{Introduction}
\label{sec:intro}
Evaluating how automated vehicles (AVs) handle ethically challenging situations in everyday driving is essential for their adoption and acceptance by society \cite{lin2016ethics, millar2017ethics}. Such situations often require trade-offs between competing values, such as safety, legality, and social norms, for which no clear or universally optimal solution exists. For instance, an AV may need to decide whether to cross a solid line to safely overtake a cyclist \cite{TeslaShort2025} or whether to come to a full stop at an empty junction when no vehicles or pedestrians are present \cite{TeslaRollingStop2025}. While human drivers usually make such choices intuitively, AVs face greater difficulty because they often depend on rule-based systems or predefined optimisation algorithms \cite{aksjonov2021rule, yuan2024evolutionary}. These systems struggle to balance safety, efficiency, regulatory compliance, and social expectations in real time, which can result in decisions that diverge from human judgement and values \cite{bin2022should}.

Addressing these dilemmas remains largely unaddressed in current AV design paradigms \cite{himmelreich2018never}. Most existing approaches to ethical decision-making focus on rare, extreme situations, such as the well-known “trolley problem” \cite{bonnefon2019trolley}. While such scenarios are philosophically intriguing, they are seldom encountered in routine driving. As Lin observes \cite{lin2016ethics}, everyday ethical challenges go well beyond rare, binary dilemmas. They require flexible, context-aware reasoning—something current AV algorithms often struggle to achieve. Similarly, Nyholm \cite{nyholm2016ethics} argues that focusing too heavily on extreme scenarios oversimplifies the probabilistic and dynamic nature of real-world driving environments.

Addressing day-to-day ethical challenges requires reasoning that considers the diverse goals of multiple human agents. In this research, we use the term human agents to include not only direct road users, such as drivers, cyclists, and pedestrians, but also those indirectly affected, including policymakers and society \cite{calvert2020conceptual}. These agents may prioritise safety, legality, efficiency, or social norms differently. As a result, ethical tensions arise when AVs must navigate between competing expectations. Recent work \cite{cecchini2024aligning, henschke2020trust} has called for more holistic approaches. Such approaches aim to integrate deontological, consequentialist, and virtue-based principles while also ensuring transparency and alignment with human moral intuitions.

However, integrating ethical principles into AV decision-making remains a challenge. Recent approaches have proposed ethical trajectory planning algorithms grounded in deontological reasoning \cite{thornton2016incorporating}, or based on risk and cost functions that combine multiple ethical considerations \cite{geisslinger2023ethical}. While these models represent progress, they have also been critiqued for lacking transparency \cite{kirchmair2023taking}. Key concerns include how ethical principles are selected, how conflicts are resolved, and how resulting decisions align with legal and societal expectations.

The principle of Meaningful Human Control (MHC) \cite{santoni2018meaningful, mecacci2020meaningful} offers a promising foundation to address these critiques. MHC is a design principle with two aims. First, AV behaviour should reflect the intentions and moral reasons of relevant human agents, a requirement known as tracking. Second, it should remain possible to assign responsibility to informed and accountable individuals, a requirement known as tracing \cite{de2023realising}. To fulfil the tracking condition, AV behaviour must be responsive to the reasons of relevant agents, including their values, plans, and intentions. It must also account for those indirectly affected, such as vulnerable road users and policymakers \cite{mecacci2020meaningful}.

MHC could serve as a conceptual bridge between ethical principles and observable AV behaviour. It links abstract moral values, such as those in deontological or utilitarian ethics, to practical elements like plans, intentions, and actions \cite{mecacci2020meaningful}. In this way, MHC suggests that moral values should be reflected in agents’ practical choices, including priorities such as safety, comfort, and rule compliance. However, before these principles can guide design, we must first evaluate whether AV decisions actually reflect them. Without a systematic evaluation method, it is impossible to judge whether an AV’s behaviour aligns with ethical expectations such as fairness, harm minimisation, or accountability. Although recent work has helped clarify the concept of MHC, the challenge remains: how can it be applied in practice to evaluate AV behaviour? This motivates the need for a framework capable of assessing whether AVs act in accordance with the moral reasons of relevant human agents.

To address this need, we propose a reason-based evaluation framework that measures how well planned AV trajectories align with the reasons of relevant human agents. Beyond trajectory alignment, the framework also tests whether an AV system satisfies the tracking condition of MHC in practice. It follows the evaluation procedure outlined by \cite{suryana2024meaningful}, which involves three steps: identifying relevant agents and their reasons, specifying the AV behaviours that should reflect those reasons, and conducting the reason evaluation. Our framework does not replace existing trajectory planning methods but evaluates their outcomes. In doing so, it provides a transparent way to determine whether a chosen trajectory aligns with the moral reasons of the agents involved.

To illustrate our approach, consider a scenario where an AV follows a slow cyclist on a road marked with double solid yellow lines, which prohibit overtaking \cite{TeslaShort2025}. After a few seconds, a human driver intervenes and overtakes, exposing a misalignment between the AV’s rule-based behaviour and human judgement. Our framework evaluates such cases by modelling the priorities of relevant agents, such as policymakers, vulnerable road users, and passengers, as mathematical functions. These functions are then used to score and compare candidate trajectories, similar to existing motion-planning pipelines. The key difference is that, instead of optimising for fixed performance criteria, we assess alignment with human reasons, providing a new layer of ethical evaluation.

Specifically, this paper introduces a novel approach for evaluating whether AV behaviour in everyday ethically challenging scenarios reflects the reasons of relevant human agents. Our primary contributions are:
\begin{enumerate}
\item We develop a \textbf{reasons-based trajectory evaluation framework} that measures the alignment between AV trajectories and the reasons of relevant human agents. This allows us to assess whether the system satisfies the tracking condition of MHC in practice.   
\item We demonstrate, through simulation, that the framework supports ethically grounded and interpretable decision-making. It does so by modelling agent influence as both quantifiable and adjustable, and by enabling both forward and inverse analysis of decisions.
\end{enumerate}

The remainder of this paper is organised as follows: Section \ref{sec:methodology} presents the methodology. Section \ref{sec:experiment} describes the experimental setup. Sections \ref{sec:results} and \ref{sec:discussion} report and discuss the results. Section \ref{sec:conclusion} concludes the paper.

\section{Methodology}\label{sec:methodology}

Current AV decision-making systems lack a mechanism to evaluate whether a selected trajectory aligns with the reasons of agents affected by it. To address this, we propose a unified trajectory scoring function that integrates agent importance, reason-level evaluations, and a fairness adjustment, thereby supporting the tracking condition of Meaningful Human Control (MHC). We begin by defining the components of the framework, then build up to the final scoring formulation.

\subsection{Human Agents and Their Reasons}

We define the human agent set \( \mathcal{H} = \{h_1, h_2, \ldots, h_n\} \), where each human \( h_i \) has a set of reasons
\[
\mathcal{R}_i = \{r_{i1}, r_{i2}, \ldots, r_{i m_i}\},
\]
with \(m_i\) denoting the number of reasons associated with human \(h_i\). Here, \(b \in \{1,\ldots,m_i\}\) indexes the individual reasons of human \(h_i\). Each human is assigned a weight \( w_i \in [0,1] \), with \( \sum_{i=1}^n w_i = 1 \). Each agent aggregates their reasons using weights \( \alpha_{ib} \in [0,1] \), where \( \sum_b \alpha_{ib} = 1 \).

\subsection{Trajectories and Environment Representation}

Given candidate trajectories \( T = \{T_1, \ldots, T_k\} \), each \(T_a \in T\) is a discretized sequence of ego states:
\[
T_a = \{s_{a0}, s_{a1}, \ldots, s_{ap}\},
\]
where \(a \in \{1,\ldots,k\}\) indexes candidate trajectories, \(p\) denotes the number of discrete time steps, and \(s_{al}\) is the ego vehicle’s state at time step \(l \in \{0,\ldots,p\}\). Each time step corresponds to \(t_l = l \cdot \Delta t\), where \(\Delta t\) is the planning time resolution. States include position, orientation, velocity, and other kinematic quantities, and are generated via feasible motion models.

In real driving, the ego vehicle’s trajectory must account for dynamic entities such as other vehicles, pedestrians, and cyclists. Since these entities influence whether human reasons can be fulfilled (e.g., safety or comfort), we include their trajectories in the evaluation.

Dynamic entities are indexed by \(q \in \{1, \ldots, Q\}\), with trajectories
\[
E_q = \{e_{q0}, \ldots, e_{qp}\},
\]
and we denote the set of all such trajectories as \( \mathcal{E} = \{E_1, \ldots, E_Q\} \). At time \(t_l\), the environment snapshot is
\[
\mathcal{E}_l = \{ e_{ql} \mid q = 1,\ldots,Q \}.
\]

\subsection{Reason-Level Evaluation}

Each reason \(r_{ib}\) has a per-time-step evaluation function
\[
f_{ib}(s_{al}, \mathcal{E}_l, t_l) : (s_{al}, \mathcal{E}_l, t_l) \rightarrow [0,1],
\]
and a trajectory-level score obtained via a temporal aggregation operator:
\begin{equation}
\label{eq3}
F_{ib}(T_a, \mathcal{E}) = \Phi\left( \{ f_{ib}(s_{al}, \mathcal{E}_l, t_l) \}_{l=0}^{p} \right).
\end{equation}

Here, \(\Phi\) maps per-time-step evaluations to a trajectory-level value. In this work, we adopt the uniform average
\begin{equation}
\Phi = \frac{1}{p+1} \sum_{l=0}^{p} (\cdot),    
\end{equation}
though alternative operators (e.g., weighted multi-objective sums \cite{xu2012real} or cumulative integral costs \cite{williams2017model}) may be used depending on design requirements or normative assumptions.

\subsection{Aggregating Reasons and Agents}

Each agent’s reason-level score is
\begin{equation}
S_i(T_a) = \sum_{b=1}^{m_i} \alpha_{ib} F_{ib}(T_a, \mathcal{E}).
\end{equation}

Combining these across agents gives the unbalanced score:
\begin{equation}
S_w(T_a) = \sum_{i=1}^{n} w_i S_i(T_a).
\end{equation}

\subsection{Agent Balance Function}

To ensure equitable agent influence and preserve MHC, we introduce an agent balance function \( B(\mathbf{w}, \mathbf{w}^*) \), which penalizes highly skewed weight configurations:

\begin{equation}
\label{eq:6}
B(\mathbf{w}, \mathbf{w}^*) = \left(1 - \frac{\sqrt{\frac{1}{n} \sum_{i=1}^{n} (w_i - w^*_i)^2}}{\sqrt{\sum_{i=1}^{n} (w^*_i)^2}}\right) \cdot \min_i \left( \frac{w_i}{w^*_i} \right)
\end{equation}
where \( \mathbf{w}^* \) is the ideal distribution, typically uniform (\( w^*_i = 1/n \)). The first term measures deviation from the ideal via RMS error, while the second ensures no agent is excluded (i.e., \( w_i > 0 \)). Together, they promote proportional fairness and representation, addressing concerns in~\cite{calvert2021gaps,mecacci2020meaningful} about agent exclusion in autonomous systems.

\subsection{Final Scoring and Trajectory Selection}

The final balanced score is
\[
S(T_a) = B(\mathbf{w}, \mathbf{w}^*) \cdot S_w(T_a).
\]

After computing \(S(T_a)\) for all trajectories, we select the one that maximises alignment with human reasons:
\[
T^* = \arg\max_{T_a \in T} S(T_a).
\]
In this work, this selection is used solely for evaluation and comparison purposes, illustrating which trajectory best aligns with human reasons.

Figure~\ref{fig:framework}, adapted from the framework structure in \cite{suryana2025iros}, illustrates how the proposed human-reasons-based evaluation module integrates into a standard hierarchical AV decision-making stack. In a conventional architecture, the global planner outputs a single nominal trajectory that is passed directly to the local controller. In our framework, we instead assume the global planner provides a set of feasible candidate trajectories. Before any trajectory reaches the controller, this candidate set is intercepted and evaluated by our human-reasons module. The module scores each candidate according to reason alignment and fairness. After scoring, the global planner then selects the best-aligned trajectory $T^*$ and returns it to the standard pipeline for execution.

In this way, our method does not alter the control architecture itself; rather, it inserts a normative evaluation layer that ensures Meaningful Human Control over the trajectory-selection stage.

\begin{figure*}
    \label{fig:framework}
    \centering
    \includegraphics[width=\linewidth]{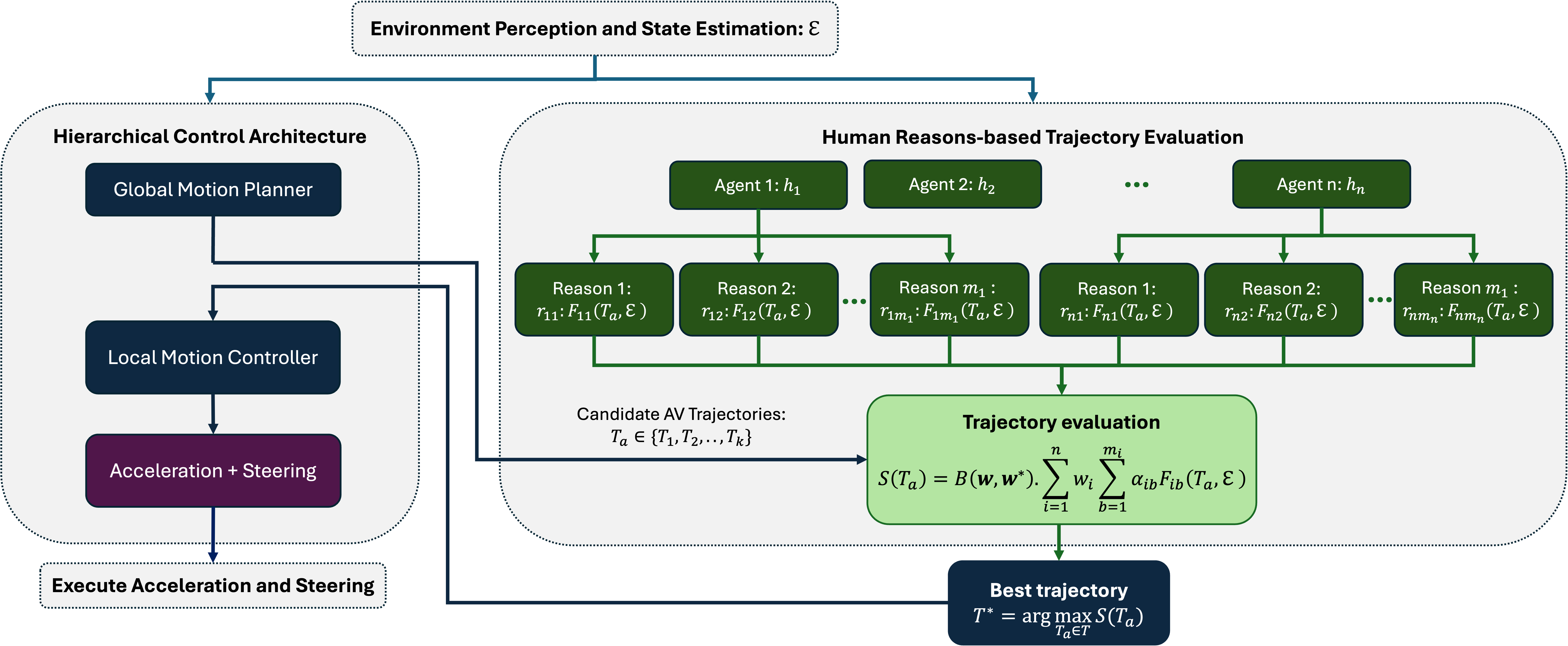}
    \caption{Integration of the proposed human-reasons-based trajectory evaluation module into a hierarchical AV control architecture. The module does not generate or select trajectories; instead, it evaluates candidate trajectories produced by the global planner for alignment with human reasons. The global planner then uses these scores to select the trajectory that best satisfies both motion-planning and human-reason considerations.}
    \label{fig:framework}
\end{figure*}

\section{Experimental Setup}
\label{sec:experiment}

\subsection{Overtaking Scenario Description}
\label{sec:overtaking_description}

To demonstrate our reasons-based trajectory evaluation framework, we implement an ethically challenging overtaking scenario involving three agents: a policymaker, a driver, and a cyclist. The scenario is adapted from a real-world case~\cite{TeslaShort2025}, where Tesla's Full Self-Driving Beta chose to remain behind a cyclist on a no-passing road, while a human driver ahead illegally overtook—highlighting tensions between safety, legality, and efficiency.

\begin{figure}[t]
\centering
\includegraphics[width=0.4\linewidth]{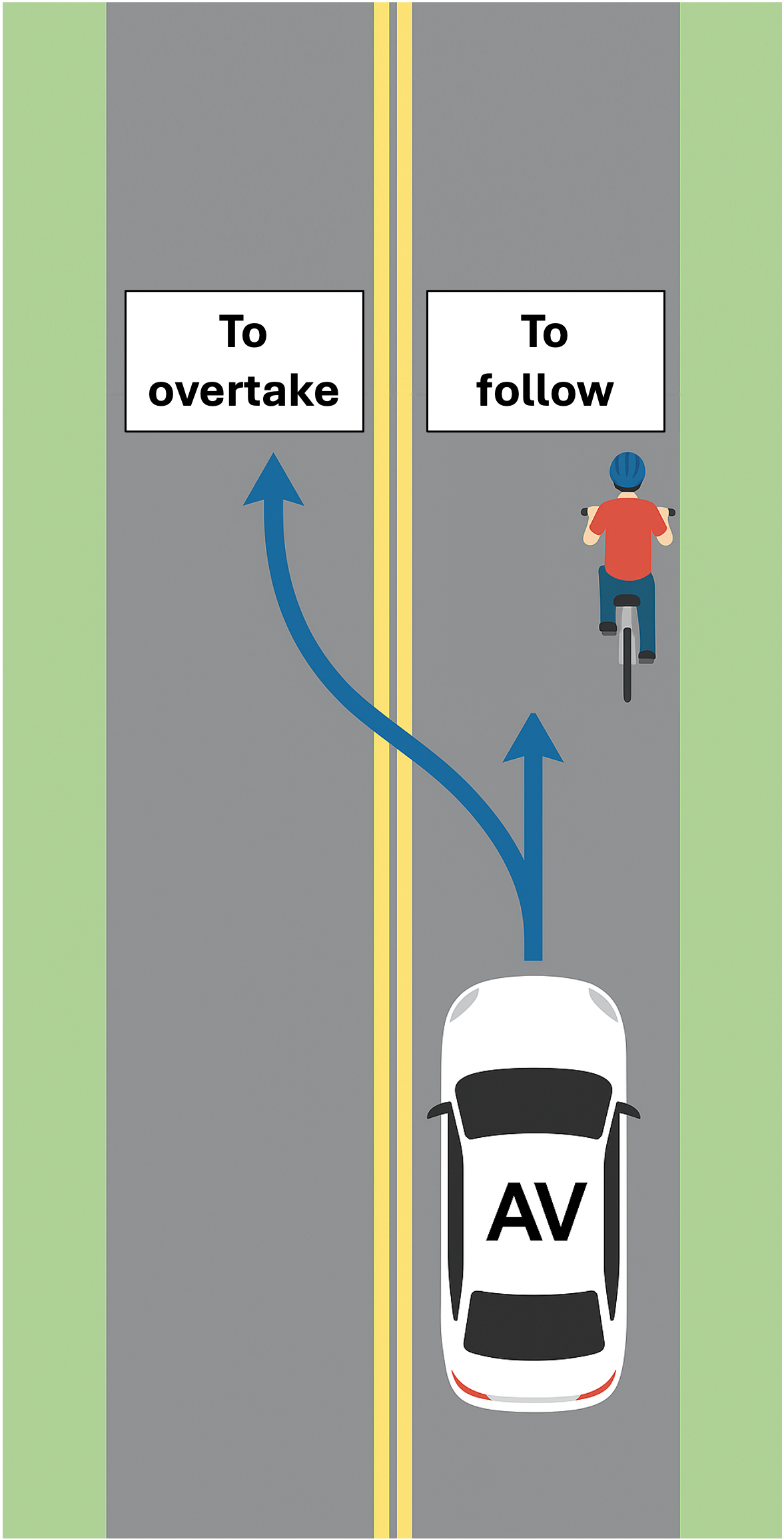}
\caption{Illustration of the vehicle-cyclist overtaking scenario showing the initial configuration and possible trajectories}
\label{fig:scenario}
\end{figure}

This situation reflects conflicts between regulatory compliance (policymaker), travel efficiency (driver), and safety/comfort (cyclist). The AV must decide whether to stay behind or overtake, trading off compliance for potential gains in efficiency. The AV encounters a slow-moving cyclist (5 km/h) on a rural two-lane road (7 m wide, 3.5 m per lane) with no oncoming traffic and a 30 km/h speed limit. A visual depiction, including the AV's trajectories, is shown in Fig.~\ref{fig:scenario}.

\subsection{Agents and Their Reasons}
\label{sec:agent_reasons}

\cite{suryana2024meaningful} evaluated safety reason alignment in partially automated driving systems using a simplified setting with two human agents and a single shared reason. While their study introduced a foundational approach to reason-based evaluation, it did not address conflicts that may arise between distinct agents with differing priorities. 

To explore such conflicts, this work models three agents, each associated with their own reason. These agents reflect a range of viewpoints commonly encountered in AV scenarios. While we focus on three agents for illustration, the framework can scale to any number of human agents, as each agent is represented as a vector \(w \in \mathbb{R}^n\).

The \textbf{policymaker} ($h_1$) prioritises regulatory compliance, such as maintaining lane discipline and ensuring the vehicle returns to the correct lane after overtaking. The \textbf{driver} ($h_2$) values time efficiency, aiming to minimise delays caused by slower vehicles while still maintaining safety. Meanwhile, the \textbf{cyclist} ($h_3$) is concerned with safety and comfort, which includes maintaining sufficient lateral clearance and expecting appropriate overtaking behaviour from surrounding vehicles. Each agent uses a single reason (\( \alpha_{i1} = 1 \)) with equal initial weight (\( w_i = 1/3 \)). We explore other weight configurations in a sensitivity analysis.

\subsection{Candidate Trajectories}\label{subsec:candidate_trajectories}

We define four candidate AV trajectories \( T = \{T_1, T_2, T_3, T_4\} \), representing different patterns of agent prioritisation in the overtaking scenario. These trajectories vary in clearance distance, lane use, and alignment with the reasons of drivers, cyclists, and policymakers. Their generation follows the procedure outlined by \cite{rahmani2023bi}, which provides a structured approach for producing AV trajectories in interaction with surrounding agents. To generate these four alternatives, we experimented with the heuristic function in the global motion planner introduced by \cite{rahmani2023bi}; however, the details of this adaptation are beyond the scope of this paper.\footnote{Code: \url{https://github.com/lucassuryana/AV-Simulation}}

\begin{figure}[h]
    \centering
    \includegraphics[width=0.3\textwidth]{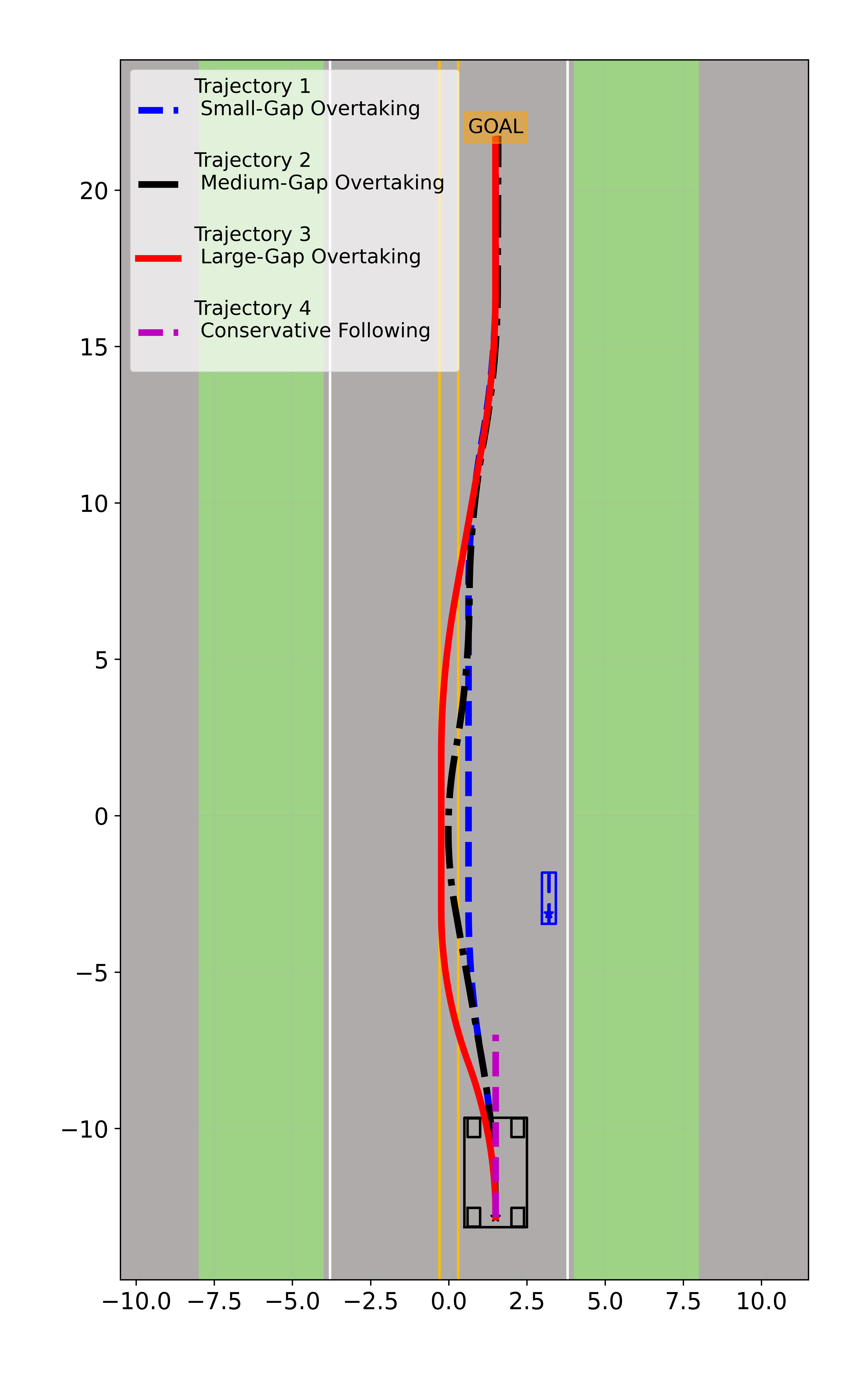}
    \caption{Spatial visualisation of four candidate AV trajectories (\(T_1\)–\(T_4\)) relative to a cyclist. The trajectories vary in lateral clearance and lane usage, reflecting different prioritisation patterns across safety, efficiency, and legal compliance.
}

    \label{fig:trajectory_spatial_view}
\end{figure}

Rather than presenting a binary decision, such as death or alive, this setup reflects the kind of everyday ethical challenges AVs are more likely to encounter—such as balancing safety, legality, and mobility. This design aligns with the critique of trolley problem framings offered by \cite{himmelreich2018never}, who advocate for a shift towards mundane driving scenarios that require context-sensitive reasoning rather than abstract moral binaries. The four trajectories are illustrated and explained in Fig.~\ref{fig:trajectory_spatial_view}. The figure illustrates four candidate trajectories:
\textbf{T\textsubscript{1}}: Small-gap overtake — minimal clearance, prioritises driver convenience, limited regard for cyclist and policymaker.
\textbf{T\textsubscript{2}}: Medium-gap overtake — balanced clearance, moderate consideration of cyclist and policymaker.
\textbf{T\textsubscript{3}}: Large-gap overtake — wide clearance, prioritises cyclist comfort and safety, least compliant with policymaker expectations.
\textbf{T\textsubscript{4}}: Conservative following — no overtake, prioritises legal compliance, lowest consideration for driver efficiency.

\subsection{Evaluation Functions and Implementation}

Each agent’s evaluation is computed via a per-time-step function \( f_{ib}(s_{al}, \mathcal{E}_l, t_l) \), introduced in Section~\ref{sec:methodology}, and averaged over the trajectory duration (Equation~\ref{eq3}). In our scenario, each agent as described in Section~\ref{sec:agent_reasons} has a single reason ($b=1$ and $\alpha_{ib}=1$); therefore, we denote reason-evaluation functions as $f_1$, $f_2$, and $f_3$ for readability.

\paragraph{Policymaker Evaluation}
Focusing on lane compliance, the policymaker's evaluation is:
\begin{equation}
f_1(s_{al}, \mathcal{E}_l, t_l) = 
\begin{cases} 
1, & d_{\text{veh}}(s_{al}) > 0, \\
e^{k_1 \cdot d_{\text{veh}}(s_{al})}, & \text{otherwise},
\end{cases}
\end{equation}
where \( d_{\text{veh}}(s_{al}) \) is the lateral distance from the lane centerline, and \( k_1 = 0.2 \) controls penalty severity.

\paragraph{Driver Evaluation}
To model time efficiency, we define a cumulative follow time \( t_{\text{elapsed},l} \) (initialized as 0). It updates each step by \( \Delta t \) if the AV is within \( d_{\text{driver}} \) of a cyclist:  
\( t_{\text{elapsed},l+1} = t_{\text{elapsed},l} + \Delta t \) if \( d_{\text{vc}} \leq d_{\text{driver}} \), else unchanged.

The driver's evaluation is:
\begin{equation}
f_2(s_{al}, \mathcal{E}_l, t_l) = 
\begin{cases} 
1, & t_{\text{elapsed},l} < t_{\text{driver}} \\ 
& \phantom{t_{\text{elapsed},l} }
\lor d_{\text{vc}} > d_{\text{driver}}, \\
\frac{1}{e^{k_2 (t_{\text{elapsed},l} - t_{\text{driver}})}}, & \text{otherwise},
\end{cases}
\end{equation}
where \( d_{\text{vc}} \) is the distance to the cyclist, and \( k_2 = 0.2 \). This formulation is supported by behavioural studies showing that driver patience declines with prolonged close following. \cite{naveteur2013impatience} link waiting time and time pressure to rising impatience. Together, these findings justify modeling satisfaction as a decaying function of follow time.

\paragraph{Cyclist Evaluation}
The cyclist's evaluation combines spatial safety and temporal comfort:
\begin{equation}
f_3(s_{al}, \mathcal{E}_l, t_l) = R_{sa}(s_{al}, \mathcal{E}_l) \cdot R_{cp}(s_{al}, \mathcal{E}_l, t_{\text{follow},l})
\end{equation}

Spatial safety component:
\begin{equation}
R_{sa}(s_{al}, \mathcal{E}_l) = 
\begin{cases} 
1, & d_{\text{vc}} > d_{\text{th}}, \\
\frac{1}{e^{k_3 (d_{\text{th}} - d_{\text{vc}})}}, & \text{otherwise},
\end{cases}
\end{equation}

Spatial temporal comfort component:  
The follow time \( t_{\text{follow},l} \) (initially 0) updates as  
\( t_{\text{follow},l+1} = t_{\text{follow},l} + \Delta t \) if \( d_{\text{vc}} \leq d_{\text{th}} \), else unchanged. The comfort score is:
\begin{equation}
R_{cp}(s_{al}, \mathcal{E}_l, t_{\text{follow},l}) = 
\begin{cases} 
1, & t_{\text{follow},l} < t_{\text{th}} \\ 
& \phantom{t_{\text{elapsed},l} }
\lor d_{\text{vc}} > d_{\text{th}}, \\
\frac{1}{e^{k_4 (t_{\text{follow},l} - t_{\text{th}})}}, & \text{otherwise},
\end{cases}
\end{equation}

Constants: \( k_3 = k_4 = 0.2 \), \( \Delta t \) is the time step, and \( d_{\text{th}} \), \( t_{\text{th}} \) are the cyclist's safety thresholds. This formulation aligns with findings from \cite{oskina2022safety}, showing that cyclists adapt behaviour—such as increasing speed and reducing lateral spacing—when followed for extended periods, indicating rising discomfort and feeling unsafe.

\subsection{Balance Function Implementation}

\begin{figure}[t]
\centering
\includegraphics[width=1.0\linewidth]{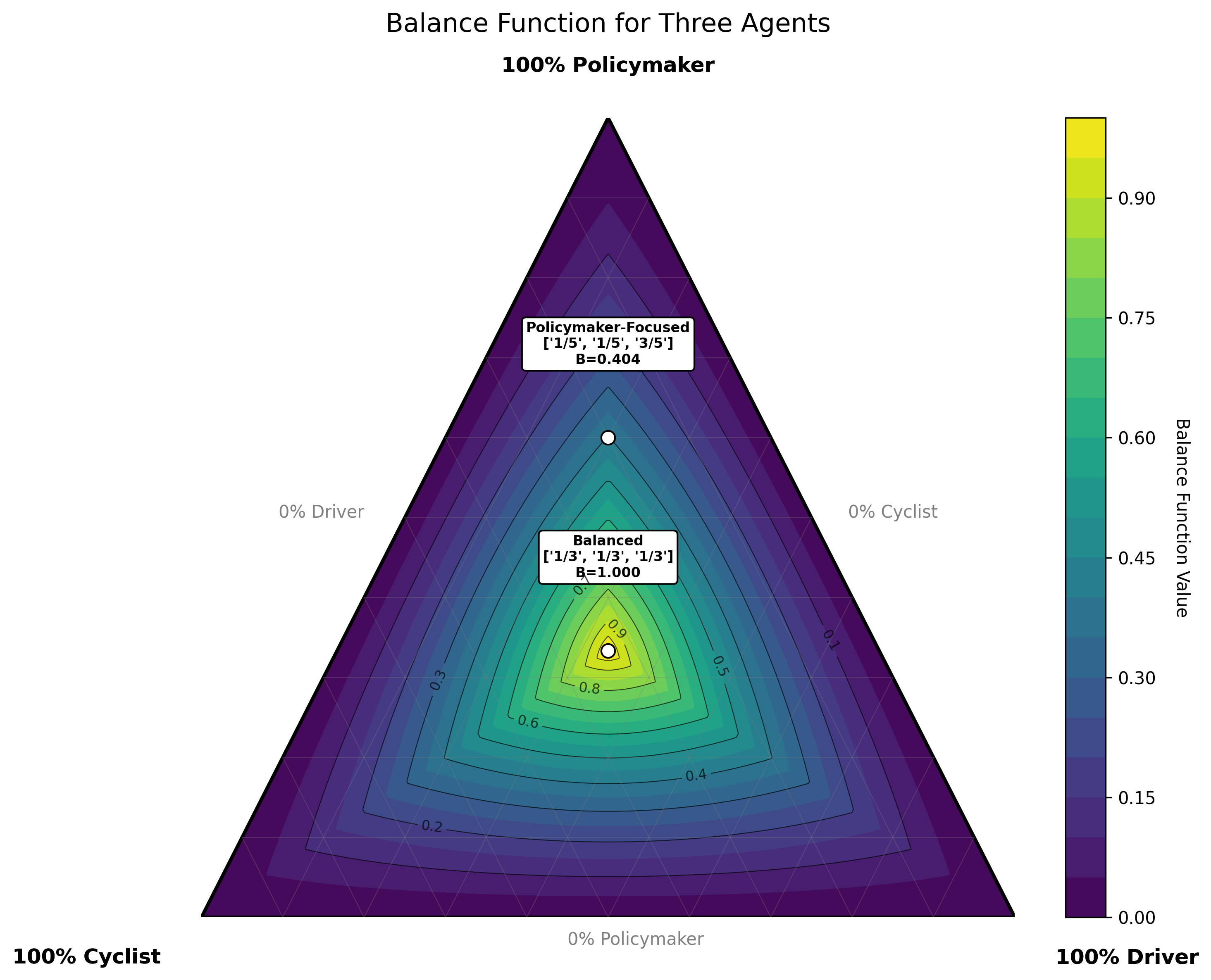}
\caption{Ternary plot showing the output of the balance function $B(\mathbf{w})$ across combinations of agent weights. Maximum balance occurs when all weights are equal.}
\label{fig:balance_plot}
\end{figure}

As per Section~\ref{sec:methodology}, the balance function \( B(\mathbf{w,w^*}) \) penalizes uneven agent weightings. For equal weights (\( w_i = 1/3 \)), \( B = 1 \); for \( w_2 = 0.6 \), \( w_1 = w_3 = 0.2 \), we get \( B = 0.487 \). Fig.~\ref{fig:balance_plot} shows the balance values across the weight simplex. The function peaks with equal influence and reaches 0 when any agent is excluded (\( w_i = 0 \)).

\section{RESULTS}\label{sec:results}
This section presents simulation results from the overtaking scenario, where each trajectory was evaluated based on its alignment with agents' reasons. Scores were computed both per agent and in aggregate using equal weighting ($w_i = 1/3$).

\begin{figure}[h]
\centering
\includegraphics[width=0.5\textwidth]{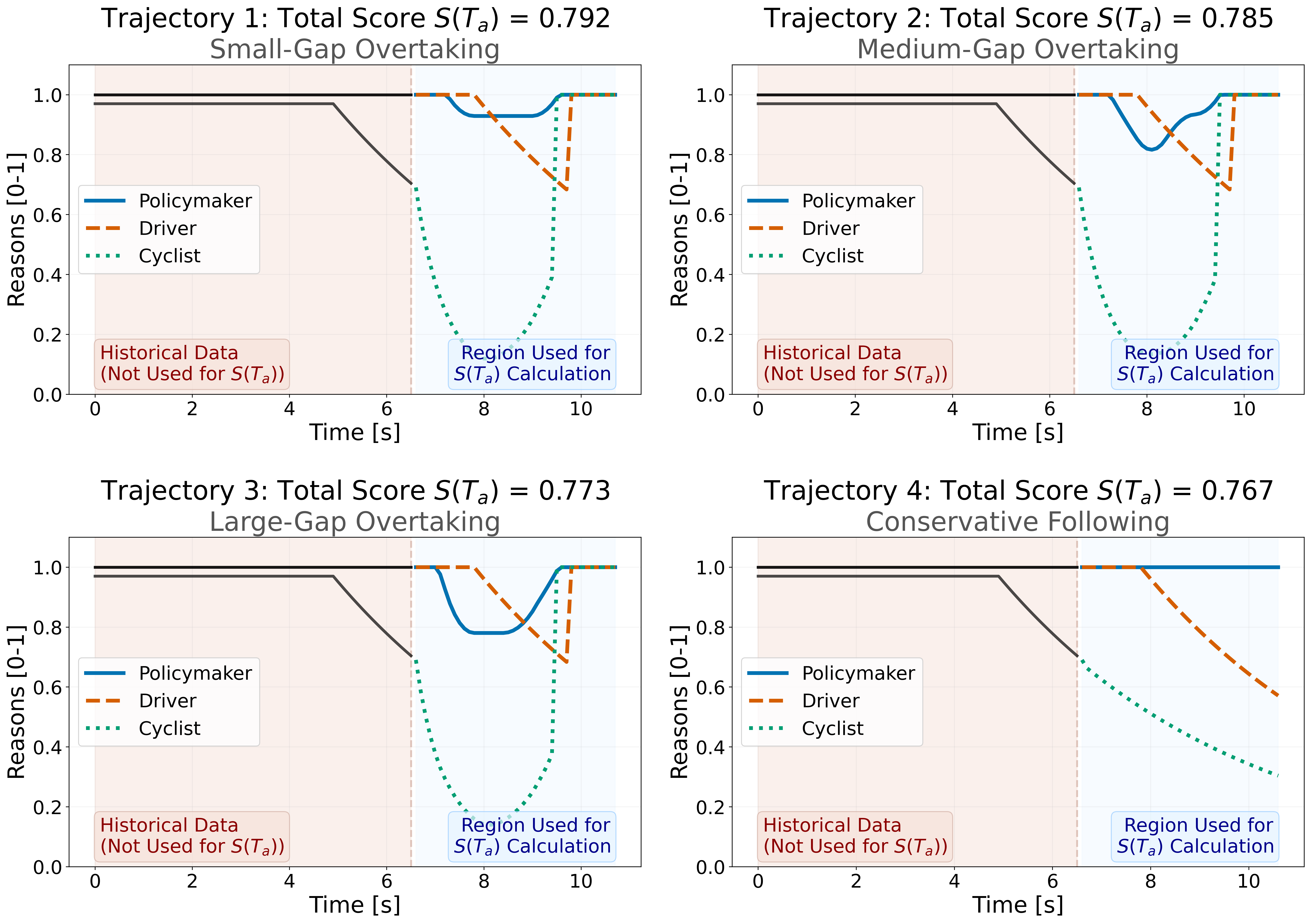}
\caption{Trajectory scores for four candidate trajectories evaluated against agents' reasons. The red region shows historical score progression; the blue region begins when the score drops below 0.7, prompting trajectory reevaluation.}
\label{fig:trajectory_scoring}
\end{figure}

We first evaluate alignment under equal agent weighting. Figure~\ref{fig:trajectory_scoring} shows the evaluation results for the four candidate trajectories. The final score $S(T_a)$ quantifies how well each trajectory aligns with the reasons of the policymaker, driver, and cyclist. The red region represents the historical progression of reason-based scores and the triggering condition for supervision, as established in previous work by \cite{suryana2025iros}. Once the score drops below the 0.7 threshold, the system generates several alternative trajectories. The blue region then begins—this marks the activation of our reason-based evaluation framework, which re-assesses the new trajectories in terms of alignment with agents’ reasons.

Among the four options, Trajectory 1 (Small-Gap Overtake) achieves the highest overall score under equal weighting, while Trajectory 4 (Conservative Following) records the lowest. This suggests that in this context, overtaking with minimal clearance better satisfies the tracking requirement across agents than remaining behind. However, trajectory rankings vary significantly depending on how agents’ importance is weighted.

\begin{figure}[h]
\centering
\includegraphics[width=0.45\textwidth]{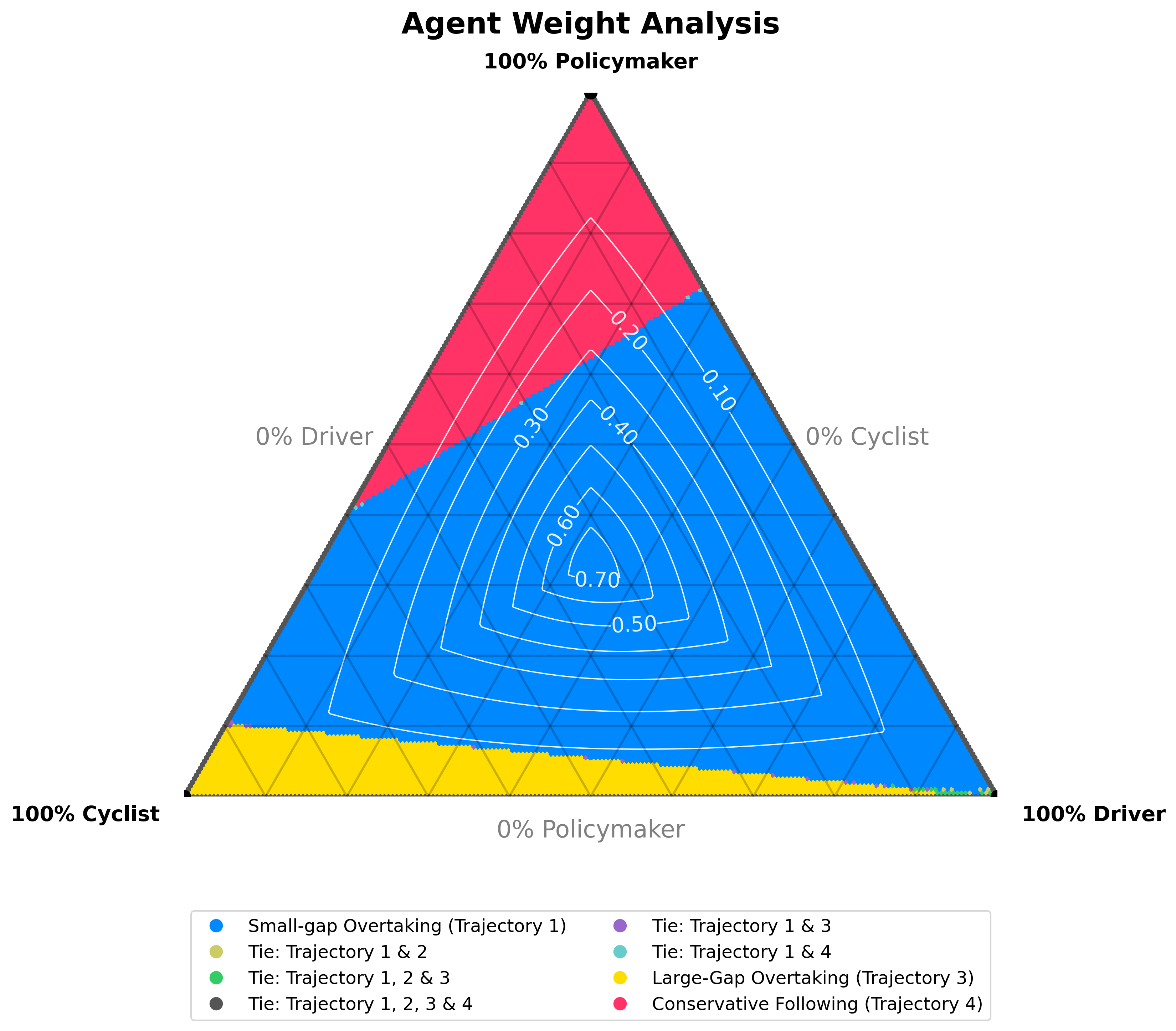}
\caption{Agent Weight Sensitivity: Optimal Trajectory Selection Across Different Priority Distributions}
\label{fig:weight_trajectories}
\end{figure}

To explore this sensitivity, we varied two agents' weights while keeping the third constant. The resulting trajectory preferences are visualised in the ternary plot in Figure~\ref{fig:weight_trajectories}, illustrating how the optimal choice depends on agent prioritisation.

Colored regions indicate which of the four trajectories achieves the highest score under each weight configuration. Blue (Trajectory 1) reflects strong driver prioritisation; Yellow (Trajectory 3) favors the cyclist; and Red (Trajectory 4) aligns with the policymaker. Other colors represent tie cases. Notably, when one agent receives zero weight (along triangle edges), all scores converge, and no clear preference emerges. White contour lines indicate score magnitudes; higher scores concentrate near regions of balanced agent influence.

These results highlight that minor shifts in agents' weights can lead to discrete changes in trajectory preference. Such critical thresholds underscore the ethical sensitivity of AV decision-making and the importance of transparent value prioritisation.

\section{Discussion}
\label{sec:discussion}

Our reasons-based evaluation framework enables automated vehicles (AVs) to assess candidate trajectories by measuring their alignment with agents' reasons. It assigns weights to each agent, computes scores, and shows how different prioritisations influence decision outcomes.

\textit{Scenario illustration and normative tension:} The simulation reflects the real-world case described in Section~\ref{sec:overtaking_description}, where strict rule-following created a misalignment between the AV’s behaviour and the reasons of relevant agents. Through weighting, the framework captures such misalignments and demonstrates how adjusted priorities can lead to alternative trajectories. These alternatives may involve short-term trade-offs but achieve closer alignment with agents’ collective reasons.

In the overtaking case, the chosen trajectory briefly enters the oncoming lane before returning. Although it achieved the highest aggregate score, it violated traffic rules and conflicted with public expectations of strict AV compliance~\cite{leenes2014laws}. Rather than endorsing such violations, the framework highlights the tensions that arise when concerns beyond regulation—such as safety and comfort—are taken into account. A similar dilemma appears when a driver mounts a kerb to let an emergency vehicle pass: technically illegal, yet often seen as serving the common good~\cite{bonnefon2020ethics}.

This example also illustrates how ethical principles surface indirectly through agents’ reasons and resulting trajectories. Prioritising safety and comfort reflects consequentialist reasoning, which focuses on outcomes. In contrast, prioritising regulation reflects deontological reasoning, which stresses rule adherence. The framework does not encode these theories directly, but their influence becomes visible through structured reasoning and trajectory evaluation.

\textit{Flexibility in prioritisation:} The overtaking case highlights one type of tension, but the framework can also accommodate AV designs that prioritise strict regulatory compliance. Giving more weight to policymakers’ reasons naturally downplays comfort and efficiency. Yet adjusting weights alone may not always change the decision. In some cases, the balance function $B(\mathbf{w}, \mathbf{w}^*)$ must also be updated so that the evaluation favours regulatory compliance. This reflects the principle of tracking in Meaningful Human Control.

As shown in the ternary plot, even small weight adjustments can trigger abrupt shifts in the selected trajectory. These threshold effects underline the importance of designing weight-setting strategies carefully and, when needed, updating the balance function to match intended design priorities.

\textit{Scalability and modular integration:} Beyond single-case illustrations, the framework is modular and can be added to existing AV motion-planning stacks. It operates as an evaluation layer over candidate trajectories, enabling the selection of the option that best balances agents’ reasons. Because it works at the evaluation layer, the framework can integrate with both modular pipelines and end-to-end learning-based planners~\cite{teng2023motion}, without requiring major changes to core control systems.

\textit{Transparency and interpretability:} A further benefit is interpretability. By quantifying agents’ reasons and assigning weights, the framework turns moral values into operational factors that directly influence trajectory selection. For example, if a chosen trajectory scores lower on regulatory compliance but higher on safety and comfort, this trade-off can be surfaced and examined.

Interpretability works in two directions. Forward interpretability checks whether trajectory selection matches predefined agent priorities. Inverse interpretability, by contrast, infers which weight configurations could have produced a given decision. Together, these support transparency by design, as proposed by~\cite{felzmann2020towards}.

This transparency could also benefit regulators. During type approval, for instance, authorities could use the framework to check whether an AV’s planned behaviour aligns with ethical expectations such as fairness and accountability~\cite{eu2018typeapproval}. They could do this without access to proprietary source code, since the framework functions as a white-box layer over decision outputs, revealing how behaviours reflect agents’ reasons.

\textit{Operationalising meaningful human control:} In addition to interpretability, the framework supports the tracking condition of Meaningful Human Control. The score function $S(T_a)$ measures how well each trajectory aligns with human reasons, while the balance function $B(\mathbf{w}, \mathbf{w}^*)$ discourages ignoring any agent. By preventing complete exclusion, the framework helps ensure that AV behaviour remains responsive to human reasons.

\textit{Limitations and future directions:} Several limitations remain. First, the framework currently assumes equal weighting across agents. While this simplifies evaluation, real-world contexts often demand unequal prioritisation—for example, stronger emphasis on safety or regulation. The balance function discourages exclusion but does not prescribe appropriate weight settings or whether they should adapt dynamically. Future work should investigate principled methods for assigning and adjusting weights.

Second, the framework assumes a correct mapping between agents’ reasons and their formal representations. This overlooks interpretive challenges in human–AV interaction. For example, regulatory compliance may be modelled as continuous, but some agents (such as law enforcement) may view it as binary. Such mismatches could undermine perceived alignment. Future studies should explore how humans interpret AV actions and whether they feel their reasons are being tracked.

Third, our evaluation focuses on a simplified overtaking scenario involving one AV and one cyclist. It does not yet capture complex planning problems, such as dense traffic, multi-agent negotiation, or long-term strategies. Extending the framework to richer scenarios would help align it more closely with real-world challenges.

Finally, future work could apply the framework to trajectories generated by different planning systems to compare their ethical alignment. Beyond AVs, the approach could be generalised to other robotic systems that rely on trajectory planning in ethically sensitive situations.

\section{Conclusion}\label{sec:conclusion}

In this work, we presented a reasons-based trajectory evaluation framework for AVs that supports decisions aligned with human agents’ reasons. The framework enables principled comparison of candidate trajectories by quantifying their alignment with agent perspectives and weighting them according to assigned priorities. Our results show that no single trajectory is universally optimal across scenarios. Instead, the best choice depends on how agent weights are configured, with different weighting schemes leading to different outcomes. This highlights the importance of carefully defining agent priorities and examining how these priorities shape AV decision-making. The framework also improves transparency by making the reasoning behind trajectory selection explicit and by supporting validation under the tracking principle of meaningful human control. Although our evaluation is simulation based, the findings demonstrate the framework’s value as a tool for assessing how AV decisions reflect agent reasons and provide a foundation for future empirical studies. Further work should investigate how to derive agent weights empirically, test the framework in real-world AV decision-making, and explore its applicability to other robotic systems that involve trajectory-based choices.

\input{report.bbl}

\bibliographystyle{IEEEtran}
% \bibliography{IEEEabrv,report}

\end{document}

%% file: report.bbl
% Generated by IEEEtran.bst, version: 1.14 (2015/08/26)